\documentclass[preprint]{elsart}
\usepackage{algorithm,algorithmic,amsmath,amssymb,framed,graphicx,multirow}
\usepackage{graphicx,amssymb}
\usepackage{algorithmic}
\usepackage{algorithm}
\usepackage{amsmath}

\usepackage{appendix}
\usepackage{longtable}
\usepackage{booktabs}

\usepackage[switch,pagewise]{lineno}
\usepackage[usenames]{color}
\usepackage{url}

\usepackage{color}

\emergencystretch=\maxdimen
\begin{document}
\begin{frontmatter}

\title{Probability Learning based Tabu Search for the Budgeted Maximum Coverage Problem}
\author[Wuhan]{Liwen Li}, \thanks{The first two authors contributed equally to this work.}
\author[Angers]{Zequn Wei}, 
\author[Angers,IUF]{Jin-Kao Hao}
\and
\author[Wuhan]{Kun He\corauthref{cor}},
\corauth[cor]{Corresponding author.} \ead{brooklet60@hust.edu.cn}


\address[Wuhan]{Huazhong University of Science and Technology, 430074 Wuhan, China}
\address[Angers]{LERIA, Universit$\acute{e}$ d'Angers, 2 Boulevard Lavoisier, 49045 Angers, France}
\address[IUF]{Institut Universitaire de France, 1 Rue Descartes, 75231 Paris, France}

\maketitle

\begin{abstract}
Knapsack problems are classic models that can formulate a wide range of applications. In this work, we deal with the Budgeted Maximum Coverage Problem (BMCP), which is a generalized 0-1 knapsack problem. Given a set of items with nonnegative weights and a set of elements with nonnegative profits, where each item is composed of a subset of elements, BMCP aims to pack a subset of items in a capacity-constrained knapsack such that the total weight of the selected items does not exceed the knapsack capacity, and the total profit of the associated elements is maximized. Note that each element is counted once even if it is covered multiple times. BMCP is closely related to the Set-Union Knapsack Problem (SUKP) that is well studied in recent years. As the counterpart problem of SUKP, however, BMCP was introduced early in 1999 but since then it has been rarely studied, especially there is no practical algorithm proposed.
By combining the reinforcement learning technique to the local search procedure, we propose a probability learning based tabu search (PLTS) algorithm for addressing this NP-hard problem. The proposed algorithm iterates through two distinct phases, namely a tabu search phase and a probability learning based perturbation phase. As there is no benchmark instances proposed in the literature, we generate 30 benchmark instances with varied properties. Experimental results demonstrate that our PLTS algorithm significantly outperforms the general CPLEX solver for solving the challenging BMCP in terms of the solution quality.

\end{abstract}

\end{frontmatter}

\section{Introduction}
\label{Sec_Intro}

In this work, we address an NP-hard problem called the Budgeted Maximum Coverage Problem (BMCP)~\cite{RefSamir1999}, which is a natural extension of the standard 0-1 Knapsack Problem (KP)~\cite{Refkellerer2003}. 
Given a set of items $\mathcal{I} = \{1, 2, ..., m\}$ where each item $i \in \mathcal{I}$ has a nonnegative weight $w_i > 0$, and a set of elements $E = \{1, 2, ..., n\}$ where each element $j \in E$ has a nonnegative profit $p_j > 0$; each item $i \in \mathcal{I}$ covers a subset of elements $E_i \subseteq E$ determined by a relationship matrix $\mathbf{M}$; given a knapsack with capacity (budget) $C > 0$; we are asked to select a subset of items $\mathcal{S} \subseteq \mathcal{I}$ such that the total weight of the selected items does not exceed the knapsack capacity and the total profit of the covered elements is maximized.  
Note that for a subset $\mathcal{S}$ of items, the profit $p_j$ of an element $j$ is counted only once in $P(\mathcal{S})$ even if the element may belong to multiple selected items.

The BMCP can be formulized as follows: 
 
\centerline{\(\text{Maximize}~~~P(\mathcal{S}) = \sum_{j \in \cup_{i\in \mathcal{S}} E_i} p_j \)}

\centerline{\(\textit{s.t.}~~~ W(\mathcal{S}) = \sum_{i \in \mathcal{S}}w_i \leqslant  C. \)}

BMCP was first proposed in 1999 and investigated in terms of approximation algorithms~\cite{RefSamir1999}, but then it is rarely studied in the literature. 
As a general problem of the NP-hard 0-1 knapsack problem, however, BMCP is not only theoretically challenging but also valuable for various domain-specific applications, such as software package installation, project assignment, database partitioning, job scheduling, facility location, etc. 

Here we provide two application scenarios. The first is on the financial decision making. The knapsack capacity corresponds to the company's project investment budget. Each item corresponds to an investment leader, together with a certain employment cost and a variety of projects to be invested. Each element corresponds to a project, together with a certain profit if invested. The goal is to hire a set of project leaders under the total budget so as to maximize the total profit of the associated projects. 
For another possible application, we provide an instance in a server storage scenario. Assume that the server has a certain storage capacity, and it needs to install some application software, each of which has a certain profit and needs to install some dependent packages in advance, which occupy a certain amount of memory space. An important decision is which set of packages should be installed to maximize the software profits without exceeding the limit of the server storage capacity. In the BMCP model of this application, each element corresponds to a software and each item corresponds to a package. The weight of each item equals the amount of memory required for the corresponding package, and the profit of each element equals the profit of each software. 

In this work, we aim to design efficient approaches for solving the BMCP in large scale, which will be the first practical algorithm for this NP-hard challenging problem. 
By combining the reinforcement learning technique with the local search procedure, we  propose a probability learning based tabu search (PLTS) algorithm. 
For this constrained binary grouping problem, there are two status for each item: selected or unselected. We use a combined neighborhood tabu-search strategy to find the local optimal solution, which is the intensification-oriented component. In order to explore more search regions, we apply a perturbation strategy based on probabilistic learning instructions as the diversification-oriented component. We associate an item with a probability vector for each possible group and determine the item group based on the probability vector. The combination of these two complementary search phases enables the algorithm to conduct a comprehensive examination of the search space. 


Our main contributions in this paper are summarized in four folds. 
First, we propose the first practical algorithm for solving the BMCP effectively. 
Second, we combines the reinforcement learning technique with the local search procedure for solving the BMCP, which is useful for solving other combinatorial optimization problems.  
Third, as there is no BMCP benchmark instances in the literature, we design and generate the first set of 30 instances with varied factors.
Forth, we compare the proposed algorithm with the general CPLEX solver and our algorithm yields competitive performance.

The rest of the paper is organized as follows. In Section 2, we provide a formal definition of the problem, followed by a review of related work in Section 3. In Section 4, we present our general algorithm framework and the composing ingredients of the probability learning based tabu search. Section 5 shows computational results and comparisons with CPLEX, and we also do ablation study to show the effectiveness of the probability learning based perturbation strategy. Section 6 concludes with a summary of major works and future research directions. 

\section{Problem Formulation}
In this section, we formulate the Budgeted  Maximum  Coverage Problem (BMCP) using a 0/1 integer linear programming model, which is also suitable for the general Integer Linear Programming (ILP) solver CPLEX. 

Given a set of items $\mathcal{I} = \{1, 2, ..., m\}$ where each item $i \in \mathcal{I}$ has a weight $w_i > 0$, and a set of elements $E = \{1, 2, ..., n\}$ where each element $j \in E$ has a profit $p_j > 0$, we are asked to select a subset of items $\mathcal{S} \subseteq \mathcal{I}$ such that the total weight of the selected items does not exceed the knapsack capacity $C$ and the total profit of their covered elements is maximized. Let $y_i~(i=1, 2, ..., m)$ be a binary variable such that $y_i = 1$ if item $i$ is selected, and $y_i = 0$ otherwise. Let $\textbf{M}$ be a $m \times n$ binary relationship matrix between $m$ items and $n$ elements where $\textbf{M}_{ij} = 1$ indicates the presence of element $j$ in item $i$. For each element $j~(j=1, 2, ..., n)$, define $H_j = \sum\limits_{i=1}^m y_i \textbf{M}_{ij}$ that counts the number of appearances of element $j$ in the items of $\mathcal{S}$. Let $x_j$ be a binary variable such that $x_j=1$ if $H_j>0$, and $x_j=0$ otherwise. 
The BMCP can be formulated as the following integer linear program.

\begin{equation*}
 \mathrm{Maximize} \quad P(\mathcal{S}) = \sum\limits_{j = 1}^n x_j p_j
\end{equation*}
\begin{equation*}
\mathrm{s.t.}~~ (1)\quad W(\mathcal{S}) = \sum\limits_{i = 1}^m y_iw_i \leqslant C
\end{equation*}
\begin{equation*}
~~~~~~~~~(2)~~y_i \in \{0,1\},~i=1,\ldots,m
\end{equation*}
\begin{equation*}
~~~~~~~~~~~~~~(3)~~H_j = \sum\limits_{i=1}^m y_i \textbf{M}_{ij},~j=1,\ldots,n
\end{equation*}
\begin{equation*}
~~~~~~(4)~~x_j = \begin{cases}
		1, & if \ H_j > 0; \\
		0, & \text{otherwise}.
		\end{cases}
\end{equation*}

\section{Related Work}
\label{realated_work}

In this section we discuss BMCP and its related problems as well as approaches for solving the related knapsack problems, and studies combining heuristics with learning techniques. 

\subsection{BMCP Related Problems}
BMCP degenerates to the NP-hard set covering problem (SCP)~\cite{RefEgonBalas1972} when the weight $w_i$, $i \in \mathcal{I}$ and profits $p_j$, $j \in E$ are all set to 1. In such a case the goal of BMCP reduces to cover as many elements as possible. BMCP can also be reduced to the standard NP-hard 0-1 knapsack problem (KP)~\cite{Refkellerer2003} when $m = n$ and each item $i \in \mathcal{I}$ covers exactly one element $i \in E$. As a generalization problem of KP and SCP, BMCP is computationally challenging. 

BMCP is closely related to  
the Set-Union Knapsack Problem (SUKP)~\cite{RefOlivier1994}. In SUKP, each item $i$ has a nonnegative profit $p_i$ and each element $j$ has a nonnegative weight $w_j$. The goal of SUKP is to package a subset $\mathcal{S}$ of items to maximize the total profit $P(\mathcal{S})$ of the selected items, while the total weight $W(\mathcal{S})$ of the covered elements does not exceed the knapsack capacity. 

SUKP can be formulized as follows:

\centerline{\(\text{Maximize}~~~P(\mathcal{S}) = \sum_{j \in \mathcal{S}}p_j \)}

\centerline{\(\textit{s.t.}~~~ W(\mathcal{S}) = \sum_{i \in \cup_{j\in \mathcal{S}} E_j} w_i \leqslant  C. \)}

\begin{figure}[h]
\centering
\label{Figure1}
\includegraphics [width=\textwidth]{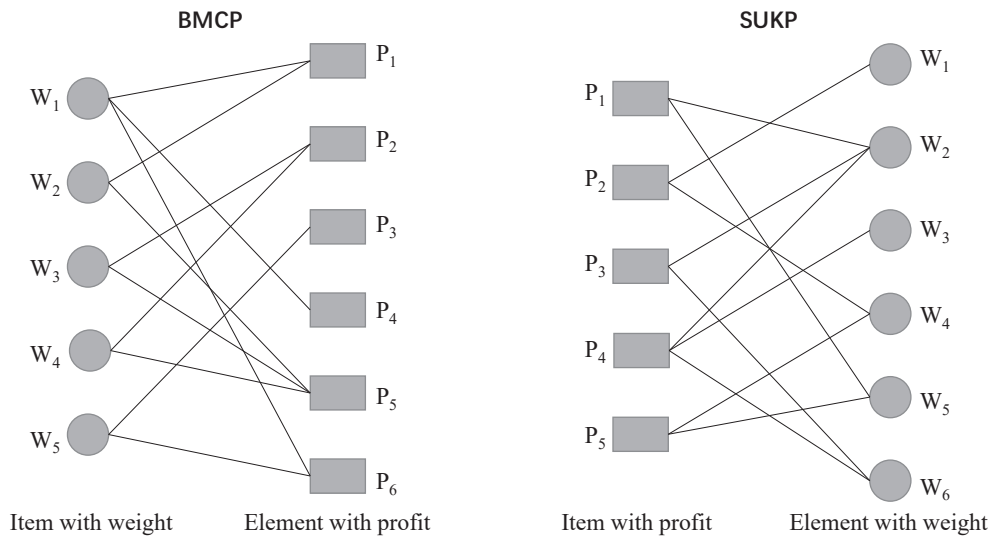}
\caption{The relationship of BMCP and SUKP.}
\end{figure}

BMCP swaps the attributes of items and elements (see Fig. 1), and thus we call BMCP the ``dual" problem of SUKP, and SUKP the ``dual" problem of BMCP. SUKP has received increasingly attention in recent years. In 1994, Goldschmidt \etal~first presented an exact algorithm based on dynamic programming to solve SUKP~\cite{RefOlivier1994}. In 2014, Arulselvan presented a greedy strategy based on an approximation algorithm~\cite{RefAshwin2014}. In 2016, Taylor designed an approximation algorithm using results of the related densest $k$-subhypergraph problem~\cite{RefTaylor2016}. Then He \etal~developed a binary artificial bee colony algorithm (BABC) for SUKP in 2018~\cite{RefYichaoHe2018}. In 2019, Wei \etal~proposed an Iterated two-phase local search I2PLS~\cite{RefZequnWei2019}, and Geng \etal~proposed a hybrid binary particle swarm optimization with tabu search~\cite{RefGenglin2019}. Very recently in 2020, He \etal~proposed a hybrid Jaya algorithm for solving SUKP~\cite{RefYichaoHe2020}.

In contrast, BMCP has attracted little attention in the literature. BMCP was first introduced by Khuller \etal~in 1999~\cite{RefSamir1999}. They also presented a $(1-\frac{1}{e})$-approximation algorithm. In 2014, Caskurlu \etal~showed that BMCP admits an $\frac{8}{9}$-approximation for bipartite graphs~\cite{RefBugra2014}. And in 2016, Kar \etal~applied BMCP in partially deployed software defined networks~\cite{RefBinayak2016}. 

\subsection{Approaches for Various Knapsack Problems}
Knapsack problems are classical NP-complete problems and well known in the field of combinatorial optimization. There are great theoretical significance and practical value for modeling and solving KPs in many fields. For approaches addressing various knapsack problems, stochastic local search has achieved considerable success in solving numerous combinatorial optimization problems~\cite{RefHoos2004}. The classic knapsack problem is the 0–1 knapsack problem (0-1 KP)~\cite{Refkellerer2003}, and many heuristic algorithms are devoted to solve various variants of the knapsack problem, e.g., multi-demand multidimensional knapsack problem~\cite{RefHalvard2006,RefLai2019}, multidimensional knapsack problem~\cite{RefKaiping2019,RefXiangjing2020}, multidimensional multiple-choice knapsack problem~\cite{RefChao2017}, quadratic knapsack problem~\cite{RefEduardo2018,RefMarcia2020} and quadratic multiple knapsack problem~\cite{RefMustafa2017,RefBingyu2018}. 

There are many members of the KPs-family, however, to our knowledge, no heuristic algorithms have been proposed for BMCP. This paper proposes a probability learning based local search algorithm for BMCP. The key part of the algorithm is inspired by the reinforcement learning, which combines the action of putting or taking out items of the knapsack with the probability vector, and realizes the search process through the perturbation strategy based on probability. This innovation method matches the integrality gap of the knapsack problem. 

\subsection{Combining Heuristics with Learning Techniques}
In recent years, researches on combining heuristics and learning techniques have received increasing attention. In 2001, Boyan and Moore proposed a learning evaluation function to improve the optimization by local search~\cite{RefJBoyan2001}. In 2016, Zhou \etal~introduced a reinforcement learning based local search (RLS) for solving the grouping problems~\cite{RefYZhou2016}. In 2017, Wang and Tang presented a machine-learning based memetic algorithm for the multi-objective permutation flowshop scheduling problem~\cite{RefXWang2017}. And Benlic \etal~proposed a hybrid breakout local search and reinforcement learning approach to the vertex separator problem~\cite{RefUnaBenlic2017}. In 2019, Jin \etal~proposed an effective reinforcement learning based local search for the maximum $k$-plex problem~\cite{RefYan2019}. And in 2020, Wang \etal~combined local search and reinforcement learning for the minimum weight independent dominating set problem~\cite{RefYiyuan2020}.

Since there do not exist polynomial time deterministic algorithms to solve KPs, in this work, we present a probability learning based local search to address the BMCP. We are interested in investigating a probabilistic guided local search method for BMCP that adopts learning technique to process information gathered from the search process so as to improve the heuristic performance.

\section{Probability Learning based Tabu Search Algorithm}
\label{Sec_Approach}
This section describes the proposed Probability Learning based Tabu Search (PLTS) algorithm for BMCP. The overall framework is introduced first, followed by the detailed algorithm description.

\subsection{General Framework}
\label{General Algorithm}
By combining the probability learning technique with tabu search, the proposed PLTS algorithm is composed of two complementary search stages: a descent-based tabu search procedure to find new local optimal solutions and a local optimal perturbation procedure based on probability learning instruction.

\begin{algorithm}
\footnotesize
\caption{Probability Learning based Tabu Search for BMCP}\label{Algo_PLTS}
\begin{algorithmic}[1]
   \STATE \textbf{Input}: Instance $A$, time limit $T_{max}$, probability vector $P$, $flip$ neighborhood $N_1$ and $swap$ neighborhood $N_2$,  tabu search depth $\alpha_{max}$
   \STATE \textbf{Output}: The best solution found $S^*$
   \STATE // Initialization of the solution $S_0$, \S \ref{Initialization} 
   \\ $S_0 \gets  Initial\_Solution(A) $
   \STATE $S^* \gets S_0$  
   \WHILE{ $RunningTime \leqslant T_{max}$}
   \STATE $P_0 \gets Initial\_Probability\_Vector(P)$
   \STATE // Optimization of the first search stage, \S \ref{Exploration} 
   \\ $(S_b, P) \gets $Tabu\_Search$(S_0, N_1,N_2, P_0, \alpha_{max}) $
   \IF{$f(S_b) > f(S^*)$}
   \STATE $S^* \gets S_b$ // Update the best solution $S^*$ found so far  
   \ENDIF
   \STATE // Optimization of the second search stage, \S \ref{perturbation} 
   \\$S_0 \gets Probability\_Perturbation(S_b, P)$
   \ENDWHILE   
   \RETURN $S^*$
\end{algorithmic}
\end{algorithm}

Specifically, the PLTS algorithm is randomly initialized and then a simple descent-based local search is applied to reach a local optimal as the initial solution. Then the tabu search procedure is adopted to explore a new local optimal solution within $flip$ neighborhood $N_1$ and $swap$ neighborhood $N_2$ (Section \ref{Neighbor}). We update the probability vector whenever a better solution is found (Section \ref{Probability}). Specifically, if an item is selected to put into the knapsack, we increase the value of the probability vector as the reward for the item. If an item is removed from the knapsack, we reduce its probability as the punishment. When the tabu search is exhausted, the PLTS algorithm uses a probability learning based perturbation to guide the search to unexplored regions. In Section \ref{perturbation}, we use the probability vector to randomly generate a new solution and start the tabu search again from this new solution. During this searching process, the best solution found is recorded and returned as the final output at the end within the time limit.

\subsection{Search Space and Evaluation Function}
\label{subsec_space}

The search space $\Omega$ explored through the tabu search process depends on the number of items in the problem instance. Given a BMCP instance composed of $m$ items $\mathcal{I} = \{1, 2, ..., m\}$ and $n$ elements $E = \{1, 2, ..., n\}$ where each item $i(i = 1, ...,m)$ corresponds to a subset of elements $E_i \subseteq E$, a candidate solution $S$ of $\Omega$ can be represented by $S=<V,\bar{V}>$ where $V$ represents the set of selected items and $\bar{V}$ represents the unselected items. The search space $\Omega$ can be represented as follows:
\begin{equation}
\Omega = \{(x_1, x_2, ..., x_m) \vert x_i \in \{0, 1\}, 1 \leqslant i \leqslant m\}.
\end{equation}

For an arbitrary solution $S \subseteq \Omega$, the total weight of $S$ is:
\begin{equation}
W(S) = \sum_{i \in S}w_i.
\end{equation} 
The objective value of the evaluation function $f(S)$ that corresponds to the total profit of $S$ is defined as:
\begin{equation}
\label{evaluation_fun}
f(S) = \sum_{j \in \cup_{i\in S} E_i} p_j.
\end{equation}

Given an instance $A$ with knapsack capacity $C$. The purpose of PLTS is to find a solution $S$ while the total weight of items $W(S) \leqslant C$ and the objective value $f(S)$ is as large as possible.

\subsection{Initialization}
\label{Initialization}
The PLTS algorithm searches from an initial solution (Algorithm \ref{Algo_Initialization}). For simplicity, we employ a simple and fast descent-based local search procedure to generate a good initial solution which is carried out in two steps. First of all, we fill the knapsack randomly until the knapsack capacity constraint is reached. Then we adopt a simple descent-based algorithm to exchange one selected item with one unselected item, which we call ``an action". If an action makes the total profit of the covered elements increase and the total weight does not exceed the limit of the knapsack capacity, then we select that move. At the end of this process, we will obtain a local optimal feasible solution as the initial solution for the tabu search procedure.

\begin{algorithm}
\footnotesize
\caption{Procedure of Generating the Initial Solution}\label{Algo_Initialization}
\begin{algorithmic}[1]
   \STATE \textbf{Function} Initial\_Solution()   
   \STATE \textbf{Input}: Instance $A$
   \STATE \textbf{Output}: An initial solution $S_0=(x_1, x_2, ..., x_m)$
   \WHILE{$TotalWeight \leqslant C$}
   \STATE Randomly add an unselected item $i$ (with weight $w_i$) into the knapsack
   \IF{$TotalWeight + w_i \leqslant C$}	
   \STATE $x_i \gets 1$
   \ELSE
   \STATE break
   \ENDIF
   \ENDWHILE
   \STATE $S_0 \gets (x_1, x_2, ..., x_m)$
   \STATE $S_0 \gets  Descent\_based\_Local\_Search(S_0)$ 
   \RETURN $S_0$
\end{algorithmic}
\end{algorithm}

\subsection{Tabu Search for Solution Improvement}
\label{Exploration}
The descent-based local search can quickly find a local optimum. However, the quality of this initial solution usually needs to be further improved. In particular, tabu search is known as one of the most popular local search methods for several knapsack problems~\cite{RefFred1997}. We take this local optimum as the input solution of the tabu search procedure (Algorithm \ref{Algo_Tabu}) to find better solution. 

The tabu search (TS) procedure examines the two neighborhoods $N_1$ and $N_2$ (Section \ref{Neighbor}) successively to explore candidate solutions. As shown in Algorithm \ref{Algo_Tabu}, at each iteration, TS picks a best neighbor solution $S' \in (N_1(S) \bigcup N_2(S))$ according to the evaluation function $f$ given by Eq.\ref{evaluation_fun} such that $S'$ is the best solution not forbidden by the tabu list. 
If no improving solution exists in $N_1(S) \bigcup N_2(S)$, the tabu search process can select the best solution $S'$ from the candidate neighborhood solutions even if $f(S') < f(S)$. This feature allows tabu search to go beyond the local optimum.

To prevent backtracking during the search, we employ a tabu list to record items involved in the swap operation. $T_i$ is the so-called tabu tenure of item $i$ and is determined as follows:
\begin{equation}
T_i = 4 + max(m, n)/100,
\end{equation} 
where $m$ is the number of items and $n$ the number of elements.
 
During the tabu search, we also need to update the probability vector of the item simultaneously 
(Section \ref{Probability}). If an item is selected into the knapsack, we will reward its probability, whereas if the item is taken out, we will reduce its probability as the punishment. This probability vector will be used during the perturbation procedure.

The tabu search process terminates when the number of iterations without improving $S'$ reaches the tabu search depth  $\alpha_{max}$. Here we apply an adaptive parameter empirically, i.e., $\alpha_{max} = (1100-m)\times 20$, so that our algorithm can automatically choose the corresponding termination conditions for different instances.

\begin{algorithm}
\footnotesize
\caption{Tabu Search Procedure}\label{Algo_Tabu}
\begin{algorithmic}[1]
    \STATE \textbf{Function} Tabu\_Search()
    \STATE \textbf{Input}: Input solution $S$, neighborhood $N_1,N_2$, probability vector $P_0$, tabu search depth $\alpha_{max}$
    \STATE \textbf{Output}: Best solution $S_b$ found during the tabu search and probability vector $P$
    \STATE $S_b \gets S$     \hfill	//$S_b$ records the best solution found so far
	\STATE $\alpha \gets 0$   \hfill //$\alpha$ counts the number of consecutive non-improving iterations  
    \WHILE {$\alpha < \alpha_{max}$}
	\STATE $S \gets argmax\{f(S'): S' \in (N_1(S) \bigcup N_2(S))\ $ and $S'$\ is\ not\ forbidden\ by\ the\ tabu\ list\}
	\STATE // Update the probability vector, \S \ref{Probability} 
	\\ $P \gets probability\_vector\_updating(P_0)$ \hfill 
    \IF    {$f(S) > f(S_b)$}
    \STATE $S_b \gets S$ \hfill // Update the best solution $S_b$ found so far
    \STATE $\alpha \gets 0$
    \ELSE  
    \STATE $\alpha \gets \alpha + 1$
    \ENDIF
    \STATE $Tabu\_list\_updating()$ 
    \ENDWHILE
    \RETURN $S_b$
\end{algorithmic}
\end{algorithm}

\subsubsection{Move Operators and Neighborhoods}
\label{Neighbor}

The neighborhood used by $Tabu\_Search()$ consists of two basic neighborhoods, namely the $flip$ 
neighborhood $N_1$ and the $swap$ neighborhood $N_2$. For a current feasible solution $S = (x_1, x_2, ..., x_m)$, the function of $flip$ neighborhood $N_1$ is to flip the value of an variable $x_q$ in $S$ while satisfying the knapsack capacity constraint $C$, that is, $Flip(q)$ changes the value of a variable $x_q$ to its complementary value $1-x_q$. Therefore, all possible solutions that can be obtained by the $filp$ operator constitute the $N_1$ neighborhood of solution $S$. $N_1(S)$ can be defined as follows:
\begin{equation}
N_1(S) = \{S' \mid S' = S \oplus Flip(q), q \in S, \sum_{i \in S}w_i \leqslant C\}.
\end{equation}

The second neighborhood $N_2$ is defined by the $swap$ operator $Swap(p, q)$ where $p$ is in the selected item set and $q$ is in the unselected item set. Note that the $swap$ operator also needs to meet the knapsack capacity constraint $C$. The $swap$ neighborhood $N_2(S)$ can be defined as follows:
\begin{equation}
N_2(S) = \{S' \mid S' = S \oplus Swap(p, q),p \in V,  q \in \bar{V}, \sum_{i \in S}w_i \leqslant C\}.
\end{equation}

The tabu search algorithm explores the union of these two neighborhoods, $N(S)=N_1(S) \bigcup N_2(S)$, and $N$ is bounded in size by $O(m + \vert V \vert \times \vert \bar{V} \vert)$.

\subsubsection{Probability Update Policy}
\label{Probability}
Our probability learning based tabu search algorithm borrows the idea of reinforcement learning in the area of machine learning. Reinforcement learning is defined as the concept of how an agent should take actions in an environment to maximize the cumulative rewards. The intuition is that actions leading to higher rewards are more likely to recur. In BMCP, for each item there are two possible states, selected or unselected. There are also two possible actions (moves) for an item: packing into the knapsack or removing from the knapsack. Since there are numerous move operations during the tabu search process, it is beneficial to integrate some learning technique to guide the search to update the probability vector.

We define a probability vector of length $n$, where $p_i$ denotes the probability that item $i$ is selected to be packed into the knapsack. Initially, all the probability values in the probability vector are set to 0.50, indicating that each item will have a half chance to be selected into the knapsack.

During the tabu search procedure, if an item $i$ is selected into the knapsack, we update its probability value as follows:
\begin{equation}
\label{reward}
p_i(t+1) = \beta + (1-\beta) \times p_i(t),
\end{equation}
where $\beta~(0<\beta<1)$ is a reward factor. On contrast, if an item $i$ is taken out of the knapsack, we punish its probability by a penalization factor $\gamma~(0<\gamma<1)$:
\begin{equation}
\label{punish}
p_i(t+1) = (1-\gamma) \times p_i(t).
\end{equation}
Our probability update scheme is inspired by the learning automata (LA)~\cite{RefNarendra1989}. The principle of this scheme is to increase the selection probability when items are packed feasibly and reduce the selection probability when items are taken out. In the tabu search procedure, the probability vector records the probability that an item is selected to pack into the knapsack. And in the perturbation procedure (Section \ref{Probability}), we can generate new solutions directly based on the probability vector.

\subsection{Probability Learning based Perturbation} 
\label{perturbation}

The purpose of the perturbation procedure is to diversify the search by exploring new search areas. The probability learning based perturbation plays an important role when the tabu search is exhausted. Specifically, each item will be dropped or picked according to the probability vector, which will generate a new perturbed solution as the starting point for the next round of tabu search. We consider and compare the following two perturbation strategies:

1) \textit{Random perturbation}: For a feasible solution, this policy randomly selects half items to be removed from the knapsack (regardless of its probability value), then uses the descent search algorithm to select items until the knapsack reaches its maximum capacity. Note that this selection policy does not use any useful information gathered from the search history.

2) \textit{Probability perturbation}: As shown in Algorithm \ref{Algo_Probability}, starting from the input local optimal solution $S_b$, this policy first drops the selected items in $S_b$ according to the probability vector $p_i$. Then, we pack unselected items into the knapsack under the guidance of $p_i$. Specifically, for a randomly unselected item $j$, we set $x_j = 1$ according to the probability vector $p_j$, when item $j$ can bring a feasible solution after being added into the knapsack. This process iterates until the knapsack capacity is reached. The new perturbed solution $S_0$ will serve as a new input solution for the tabu search. Thus, the probability perturbation makes fully utilization of the probability vector. If an item has a high probability of being selected, it has a higher probability of being taken out of the knapsack. On the contrary, if the value of probability vector is small, it has a higher probability of being selected into the knapsack. This strategy enables the algorithm to explore new search areas from a feasible solution. Furthermore, we will show the impact of this perturbation strategy on the algorithm performance in Section \ref{Analysis_Sec}.


\begin{algorithm}
\footnotesize
\caption{Probability Perturbation Policy}\label{Algo_Probability}
\begin{algorithmic}[1]
   \STATE \textbf{Function} Probability\_Perturbation()   
   \STATE \textbf{Input}: Input solution $S_b$, number of items $(m)$, probability vector $P$, knapsack capacity limit $C$
   \STATE \textbf{Output}: New solution $S_0=(x_1, x_2, ..., x_m)$
   \FOR{each selected item $i$ in $S_b$ ($x_i = 1$)}  
   \STATE $p \gets rand(0, 1)$
   \IF{$p \leqslant p_i$} 
   \STATE $x_i \gets 0$   \hfill  //Drop item according to probability vector
   \STATE Update $TotalWeight$
   \ENDIF
   \ENDFOR 
   \FOR{each unselected item $j$ in $S_b$ ($x_j = 0$)}  
   \IF{$TotalWeight + w_j \leqslant C$}
   \STATE $p \gets rand(0, 1)$
   \IF{$p > p_j$} 
   \STATE $x_j \gets 1$  \hfill //Pick item according to probability vector
   \STATE Update $TotalWeight$
   \ENDIF
   \ELSE
    \STATE \textbf{break}
   \ENDIF
   \ENDFOR 
   \STATE $S_0 \gets (x_1, x_2, ..., x_m)$
   \RETURN $S_0$
\end{algorithmic}
\end{algorithm}

\section{Experimental Results}
\label{Sec_Results}

In this section, we present the experimental results of the proposed PLTS algorithm on 30 benchmark instances that we designed for BMCP, and show comparisons with the results obtained by the CPLEX slover. Then we analyze the probability learning based perturbation of the PLTS algorithm.


\subsection{Benchmark Instances}
\label{Instance}

As there are no existing benchmark instances for BMCP, inspired by the instances of SUKP~\cite{RefYichaoHe2018}, we generate 30 instances with similar characteristics to the instances of SUKP. These instances are divided into three sets, ranging from 585 to 1000, based on the relationship between the number of items and the number of elements~\footnote{Our BMCP dataset: The link will be available after publication.}. The number of items in the first group is less than the number of elements. In the second group the number of items equals the number of elements and in the third group the number of items is greater than the number of elements. Let $\textbf{M}$ be a $m \times n$ binary relationship matrix between $m$ items and $n$ elements where $\textbf{M}_{ij} = 1$ indicates the presence of element $j$ in item $i$. To avoid the number selection of items that easily cover all elements, we adjusted the density of the relationship matrix $\textbf{M}$ to a fixed value according to the capacity of the knapsack. When the knapsack capacity is 2000, the density of the relationship matrix is 0.05, and when the knapsack capacity is 1500, the density of the relationship matrix is 0.075. Then $bmcp\_m\_n\_\alpha\_C$ designates an instance with $m$ items and $n$ elements, density of relationship matrix $\alpha$ and knapsack capacity $C$, where $\alpha = (\sum_{i=1}^m \sum_{j=1}^n M_{ij})/(mn)$. Experimental results of the three sets of instances are shown in Tables 2-4.

\subsection{Parameter Setup}
\label{Setting}
The proposed PLTS algorithm was coded in C++ and and compiled using the g++ compiler with the -O3 option. The experiments were carried on an Intel Xeon E5-2670 processor with 2.5 GHz and 2 GB RAM under the Linux operating system.

\renewcommand{\baselinestretch}{0.9}\large\normalsize
\begin{table}[h]\centering
\begin{scriptsize}
\caption{Parameter setup.}
\label{Parameter_Settings}
\begin{tabular}{cclc}
\hline
Parameters	   &	Section				    & Description 		    & Value	\\
\hline
$T_{max}$	   &	\ref{General Algorithm}	& time limit            & 600	\\
$\beta$        &    \ref{Probability}       & reward factor 		& 0.50	\\
$\gamma$       &    \ref{Probability}	    & penalization factor	& 0.50	\\
\hline
\end{tabular}
\end{scriptsize}
\end{table}
\renewcommand{\baselinestretch}{1.0}\large\normalsize

Table \ref{Parameter_Settings} shows the description and setting of the parameters used for experiments. For the sake of fairness and convenience, we set both the reward factor and the punishment factor as 0.50 respectively. To obtain the experimental results, each instance was solved 30 times independently with different random seeds, and the cut-off time is set as 600 seconds per run. 

Since there is no result reported by using the general integer linear programming (ILP) approach for solving the BMCP, we present computational results attained by the CPLEX solver (version 12.8) under a time limit of 2 hours for each instance based on the 0/1 integer linear programming model. The CPLEX solver finds the upper bound and lower bound of the instances and the experimental results are shown in Tables 2-4.

\subsection{Comparison on Computational Results}
\label{Result}
We first assess the performance of the proposed PLTS algorithm with respect to the CPLEX solver. In Table 2-4 we report the computational results of PLTS and CPLEX on the three sets of benchmark instances. Here the first column shows each instance name, followed by the best lower bound (LB) and upper bound (UB) achieved by CPLEX; then, $f_{best}$ indicates the best objective value obtained by PLTS over 30 runs, followed by the average value ($f_{avg}$), standard deviations ($Std$), and average running time ($t_{avg}$) in seconds.

\begin{table}[!htp]\centering
	\caption{Comparison of PLTS with the CPLEX solver on the first set of instances ($m<n$).} 
	\renewcommand{\baselinestretch}{0.9}\large\normalsize
	\begin{scriptsize}
	\setlength{\tabcolsep}{1mm}{
	\begin{tabular}{lcc|cccc}
	\toprule[0.75pt]

\multirow{2}{*}{Instance} & \multicolumn{2}{c|}{CPLEX} & \multicolumn{4}{c}{PLTS} \\
\cline{2-7}
    &  LB  & UB & $f_{best}$ & $f_{avg}$ & $Std$ & $t_{avg}$  \\

\hline
bmcp\_585\_600\_0.05\_2000  & 67910 & 73495.88    & \textbf{71102}  & 71065.17    & 82.36    & 309.602  \\
bmcp\_585\_600\_0.075\_1500 & 68418 & 77549.43    & \textbf{70677}  & 70677.00    & 0.00     & 61.242   \\
bmcp\_685\_700\_0.05\_2000  & 79997 & 88954.29    & \textbf{81227}  & 80585.73    & 508.37   & 522.060  \\  
bmcp\_685\_700\_0.075\_1500 & 80443 & 92328.30    & \textbf{82955}  & 82951.40    & 19.39    & 109.670  \\
bmcp\_785\_800\_0.05\_2000  & 90705 & 102198.73   & \textbf{92608}  & 92587.60    & 34.30    & 252.589  \\
bmcp\_785\_800\_0.075\_1500 & 92358 & 107354.51   & \textbf{94245}  & 94245.00    & 0.00     & 248.128  \\
bmcp\_885\_900\_0.05\_2000  & 100085 & 114313.81  & \textbf{102162} & 101331.53   & 174.95   & 206.025  \\
bmcp\_885\_900\_0.075\_1500 & 102423 & 122684.38  & \textbf{106577} & 105942.43   & 334.18   & 489.396  \\
bmcp\_985\_1000\_0.05\_2000 & 107820 & 125903.67  & \textbf{109567} & 109408.77   & 227.85   & 212.193  \\
bmcp\_985\_1000\_0.075\_1500 & 110769 & 134440.36 & \textbf{114969} & 113838.07   & 509.34   & 485.677  \\

	\bottomrule[0.75pt]
	\end{tabular}
	}
\end{scriptsize}
\label{Results1}
\end{table}
\renewcommand{\baselinestretch}{1.0}\large\normalsize

\begin{table}[!htp]\centering
	\caption{Comparison of PLTS with the CPLEX solver on the second set of instances ($m=n$).} 
	\renewcommand{\baselinestretch}{0.9}\large\normalsize
	\begin{scriptsize}
	\setlength{\tabcolsep}{1mm}{
	\begin{tabular}{lcc|cccc}
	\toprule[0.75pt]
	
\multirow{2}{*}{Instance} & \multicolumn{2}{c|}{CPLEX} & \multicolumn{4}{c}{PLTS} \\
\cline{2-7}
    &  LB  & UB & $f_{best}$ & $f_{avg}$ & $Std$ & $t_{avg}$  \\

\hline
bmcp\_600\_600\_0.05\_2000    & 67917  & 73495.08  & \textbf{68738}  & 68472.00   & 71.09    & 95.638  \\   
bmcp\_600\_600\_0.075\_1500   & 70947  & 77379.97  & \textbf{71746}  & 71746.00   & 0.00     & 27.975  \\
bmcp\_700\_700\_0.05\_2000    & 76367  & 85587.33  & \textbf{78028}  & 77859.27   & 75.16    & 127.445 \\
bmcp\_700\_700\_0.075\_1500   & 81645  & 93026.66  & \textbf{84576}  & 84375.70   & 550.91   & 196.995 \\
bmcp\_800\_800\_0.05\_2000    & 90344  & 101141.03 & \textbf{91795}  & 91576.27   & 309.05   & 307.274 \\
bmcp\_800\_800\_0.075\_1500   & 94049  & 108713.00 & \textbf{95533}  & 95509.60   & 70.20    & 239.146 \\
bmcp\_900\_900\_0.05\_2000    & 100108 & 115682.55 & \textbf{101265} & 101231.17  & 62.94    & 325.683 \\
bmcp\_900\_900\_0.075\_1500   & 101035 & 120452.83 & \textbf{104521} & 104521.00  & 0.00     & 176.865 \\
bmcp\_1000\_1000\_0.05\_2000  & 109928 & 131194.18 & \textbf{112802} & 111897.07  & 636.78   & 577.668 \\
bmcp\_1000\_1000\_0.075\_1500 & 115313 & 139152.89 & \textbf{120246} & 118467.87  & 546.67   & 279.220 \\

	\bottomrule[0.75pt]
	\end{tabular}
	}
\end{scriptsize}
\label{Results2}
\end{table}
\renewcommand{\baselinestretch}{1.0}\large\normalsize

\begin{table}[!htp]\centering
	\caption{Comparison of PLTS with the CPLEX solver on the third set of instances ($m>n$).} 
	\renewcommand{\baselinestretch}{0.9}\large\normalsize
	\begin{scriptsize}
	\setlength{\tabcolsep}{1mm}{
	\begin{tabular}{lcc|cccc}
	\toprule[0.75pt]
	
\multirow{2}{*}{Instance} & \multicolumn{2}{c|}{CPLEX} & \multicolumn{4}{c}{PLTS} \\
\cline{2-7}
    &  LB  & UB & $f_{best}$ & $f_{avg}$ & $Std$ & $t_{avg}$  \\

\hline
bmcp\_600\_585\_0.05\_2000   & 66184  & 71739.68  & \textbf{67636}  & 67460.80   & 350.40   & 202.660  \\
bmcp\_600\_585\_0.075\_1500  & 68145  & 77321.51  & \textbf{70588}  & 70406.63   & 105.09   & 584.205  \\
bmcp\_700\_685\_0.05\_2000   & 76139  & 83561.86  & \textbf{78054}  & 78037.00   & 51.00    & 197.590  \\
bmcp\_700\_685\_0.075\_1500  & 75841  & 86956.88  & \textbf{78869}  & 78869.00   & 0.00     & 46.987   \\
bmcp\_800\_785\_0.05\_2000   & 86139  & 97891.07  & \textbf{89138}  & 88581.20   & 103.40   & 204.141  \\
bmcp\_800\_785\_0.075\_1500  & 89018  & 103644.12 & \textbf{91021}  & 91010.20   & 25.67    & 297.211  \\
bmcp\_900\_885\_0.05\_2000   & 97088  & 110828.83 & \textbf{98840}  & 98718.00   & 151.34   & 227.976  \\
bmcp\_900\_885\_0.075\_1500  & 99881  & 119045.52 & \textbf{105141} & 104397.93  & 691.61   & 229.644  \\
bmcp\_1000\_985\_0.05\_2000  & 108714 & 126779.11 & \textbf{111859} & 111228.80  & 828.72   & 244.920  \\
bmcp\_1000\_985\_0.075\_1500 & 108801 & 131873.01 & \textbf{112250} & 112125.87  & 143.22   & 234.614  \\

	\bottomrule[0.75pt]
	\end{tabular}
	}
\end{scriptsize}
\label{Results3}
\end{table}
\renewcommand{\baselinestretch}{1.0}\large\normalsize

\begin{figure}
\centering
\label{Figure2}
\includegraphics [width=1\textwidth]{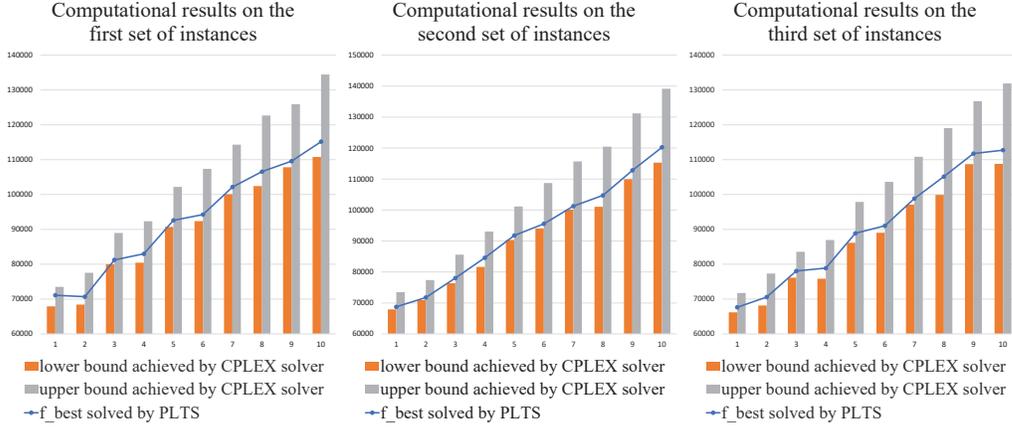}
\caption{Comparison on results of PLTS with upper and lower bounds of CPLEX on the three sets of instances.}
\end{figure}


In Figure 2, we further illustrate the comparative results of PLTS and CPLEX. The histogram shows the upper and lower bounds obtained by the CPLEX solver, and the line graph shows the best solutions obtained by the PLTS algorithm. Here X-axis represents the corresponding 10 instances for each data set respectively. Figure 2 indicates that the proposed PLTS algorithm significantly outperforms the CPLEX solver. For all the 30 instances, the best solutions ($f_{best}$) obtained by PLTS are larger than the lower bound (LB) obtained by the CPLEX solver. This study shows that the tabu search procedure and the probability learning based procedure of PLTS significantly boosts the algorithm performance for solving the BMCP.

\subsection{Ablation Study on Perturbation Policy}
\label{Analysis_Sec}
We further analyze the main ingredients of the PLTS algorithm, the probability learning based perturbation. In Section \ref{perturbation}, we present two strategies to escape from the local optimal solution, random perturbation and probability perturbation. Here we compare these two perturbation strategies, which allows us to better understand the behavior of PLTS and shed light on its inner functioning.

\begin{table}[!htp]\centering
	\caption{Comparison of PLTS with PLTS$_0$ on the three sets of BMCP instances.} 
	\renewcommand{\baselinestretch}{0.9}\large\normalsize
	\begin{scriptsize}
	\setlength{\tabcolsep}{1mm}{
	\begin{tabular}{lcc|cc}
	\toprule[0.75pt]

\multirow{2}{*}{Instance} & \multicolumn{2}{c|}{PLTS} &  \multicolumn{2}{c}{PLTS$_0$} \\
\cline{2-5}
    &  $f_{best}$  & $f_{avg}$ &  $f_{best}$  & $f_{avg}$ \\

\hline
bmcp\_585\_600\_0.05\_2000  & 71102          	& \textbf{71065.17}  & 71102   & 71056.97         \\
bmcp\_585\_600\_0.075\_1500 & 70677          	& 70677.00     	     & 70677   & 70677.00         \\
bmcp\_685\_700\_0.05\_2000  & 81227         	& \textbf{80585.73}	 & 81227   & 80578.67         \\  
bmcp\_685\_700\_0.075\_1500 & 82955			 	& \textbf{82951.40}  & 82955   & 82947.80         \\
bmcp\_785\_800\_0.05\_2000  & \textbf{92608}    & \textbf{92587.60}  & 92599   & 92492.20         \\
bmcp\_785\_800\_0.075\_1500 & 94245        	 	& \textbf{94245.00}  & 94245   & 94244.60         \\
bmcp\_885\_900\_0.05\_2000  & \textbf{102162}	& \textbf{101331.53} & 101834  & 101259.73        \\
bmcp\_885\_900\_0.075\_1500 & 106577			& 105942.43  & \textbf{106723} & \textbf{106056.07} \\
bmcp\_985\_1000\_0.05\_2000 & \textbf{109567}	& \textbf{109408.77} & 109470  & 109042.17        \\
bmcp\_985\_1000\_0.075\_1500 & \textbf{114969}	& \textbf{113838.07} & 114567  & 113583.03        \\
bmcp\_600\_600\_0.05\_2000  & 68738     		& 68472.00           & 68738   & \textbf{68488.60}  \\
bmcp\_600\_600\_0.075\_1500 & 71746        		& 71746.00           & 71746   & 71746.00         \\
bmcp\_700\_700\_0.05\_2000  & \textbf{78028}    & 77859.27           & 77910   & \textbf{77880.17}  \\  
bmcp\_700\_700\_0.075\_1500 & 84576     		& \textbf{84375.70}  & 84576   & 84257.03         \\
bmcp\_800\_800\_0.05\_2000  & 91795     		& \textbf{91576.27}  & 91795   & 91392.07         \\
bmcp\_800\_800\_0.075\_1500 & 95533     		& \textbf{95509.60}  & 95533   & 95500.73         \\
bmcp\_900\_900\_0.05\_2000  & 101265   			& \textbf{101231.17} & 101265  & 101165.93        \\
bmcp\_900\_900\_0.075\_1500 & 104521		    & 104521.00          & 104521  & 104521.00        \\
bmcp\_1000\_1000\_0.05\_2000 & \textbf{112802}  & \textbf{111897.07} & 112597  & 111560.43        \\
bmcp\_1000\_1000\_0.075\_1500 & \textbf{120246} & \textbf{118467.87} & 119533  & 118453.60        \\
bmcp\_600\_585\_0.05\_2000  & 67636       		& \textbf{67460.80}  & 67636   & 67373.20         \\
bmcp\_600\_585\_0.075\_1500 & 70588    			& \textbf{70406.63}  & 70588   & 70357.63         \\
bmcp\_700\_685\_0.05\_2000  & 78054       		& \textbf{78037.00}  & 78054   & 77992.67         \\  
bmcp\_700\_685\_0.075\_1500 & 78869         	& 78869.00           & 78869   & 78869.00         \\
bmcp\_800\_785\_0.05\_2000  & \textbf{89138}    & 88581.20           & 89084   & \textbf{88611.63}  \\
bmcp\_800\_785\_0.075\_1500 & 91021      		& \textbf{91010.20}  & 91021   & 91006.70         \\
bmcp\_900\_885\_0.05\_2000  & 98840       		& 98718.00           & 98840   & \textbf{98747.57}  \\
bmcp\_900\_885\_0.075\_1500 & \textbf{105141}   & \textbf{104397.93} & 105076  & 104253.67        \\
bmcp\_1000\_985\_0.05\_2000 & \textbf{111859} 	& \textbf{111228.80} & 111802  & 111154.23        \\
bmcp\_1000\_985\_0.075\_1500 & 112250			& \textbf{112125.87} & 112250  & 111885.13        \\
\hline
$p$-$value$ & - & - & 2.08e-5 & 2.62e-3 \\

	\bottomrule[0.75pt]
	\end{tabular}
	}
\end{scriptsize}
\label{Results4}
\end{table}
\renewcommand{\baselinestretch}{1.0}\large\normalsize

To verify the effectiveness of the probability learning based perturbation used in PLTS, we made a comparison between the probability perturbation and the random perturbation, in which we removed the probability learning mechanism from the PLTS algorithm.
Denote the modified algorithm using the random perturbation as PLTS$_0$.

The investigation was conducted on the same sets of instances we generated and each algorithm was run 30 times to solve each instance. The comparative results between PLTS and PLTS$_0$ are summarized in Table \ref{Results4}. For each instance, we report the best solution ($f_{best}$) and average solution ($f_{avg}$) of each algorithm, and better results (with a larger $f_{best}$ or $f_{avg}$) between the two are in bold. The $p$-$values$ from the Wilcoxon signed rank test are reported in the last row.

As shown in Table \ref{Results4}, the perturbation strategy has a significant impact on the performance of our algorithm. PLTS improves on the best-known results for 10 out of 30 instances compared to the random perturbation. There are 19 instances where the two methods yield same results, and only in one case the random perturbation achieves better result. As for the average solution, there are 21 out of 30 instances that PLTS yields better results. Only 5 instances PLTS$_0$ yields better solutions and 4 instances the two methods yield same results. Moreover, the $p$-$values$ ($< 0.05$) from the Wilcoxon signed rank test disclose that the difference between PLTS and PLTS$_0$ is significant. These indicate that the probability perturbation strategy with probability vector is significantly better than the simple random perturbation algorithm.

\section{Conclusions}
\label{conclusions}
In this paper, we presented the probability learning based tabu search algorithm for solving the Budgeted Maximum Coverage Problem (BMCP), which is a generalization of the 0-1 knapsack problem and a dual problem of the popular Set-Union Knapsack Problem (SUKP). BMCP is valuable for a variety of practical applications, but it has received little attention in the literature. The proposed PLTS algorithm combines probability learning techniques and a tabu search procedure within the iterated local search framework. Probability learning is used to maintain and update a probability vector, with each entry specifying the probability that the corresponding item is selected. At each iteration, the PLTS algorithm applies a tabu search procedure to improve the solution until a local optimum is reached, then PLTS adopts a perturbation phase based on the probability vector to escape from the local trap. 

As there exists no benchmark instances for BMCP, we generated three sets with a total of 30 instances and compare the results of PLTS with CPLEX. For all the 30 instances, the best solutions obtained by PLTS were larger than the lower bound (LB) obtained by the CPLEX solver. We also showed an ablation study on the probability learning based perturbation and made a comparison between probability perturbation and random perturbation, which shows that probability perturbation is much more effective than the random perturbation for solving BMCP. 

The way of using probability learning method in PLTS is innovative. Still there is room for further improvement. In future work, it is worth testing other local search algorithms or apply other heuristics for solving BMCP. Also, our probability learning based tabu search  approach could be applied to other knapsack problems or other combinatorial optimization problems, such as multidimensional knapsack problems and multiple knapsack problems.

\bibliographystyle{plain} 

\bibliography{mybibfile}


\end{document}